\documentclass[11pt,a4paper,dvipsnames]{article}
\usepackage[hyperref]{acl2021}
\usepackage{times}
\usepackage{latexsym}

\usepackage{graphicx} % For images
\usepackage{tabularx} % For tables
\usepackage{float} % Also for tables
\usepackage{enumitem} % For bulleted lists
\usepackage{xcolor}
\usepackage{booktabs}
\usepackage{multicol}
\usepackage[utf8x]{inputenc} 

\usepackage{microtype}

\usepackage{amsmath}
\newcommand{\argminB}{\mathop{\mathrm{argmin}}} 

\aclfinalcopy % Uncomment this line for the final submission
\def\aclpaperid{2739} %  Enter the acl Paper ID here

\title{Sentiment-based Candidate Selection For NMT}

 \author{Alex Jones \\
   Dartmouth College \\
   \texttt{alexander.g.jones.23@dartmouth.edu} \\\And
   Derry Wijaya \\
   Boston University \\
   \texttt{wijaya@bu.edu} \\}

\date{}

\begin{document}
\maketitle
\begin{abstract}
The explosion of user-generated content (UGC)—e.g. social media posts, comments, and reviews—has motivated the development of NLP applications tailored to these types of informal texts. Prevalent among these applications have been sentiment analysis and machine translation (MT). Grounded in the observation that UGC features highly idiomatic, sentiment-charged language, we propose a decoder-side approach that incorporates automatic sentiment scoring into the MT candidate selection process. We train separate English and Spanish sentiment classifiers, then, using n-best candidates generated by a baseline MT model with beam search, select the candidate that minimizes the absolute difference between the sentiment score of the source sentence and that of the translation, and perform a human evaluation to assess the produced translations. Unlike previous work, we select this minimally divergent translation by considering the sentiment scores of the source sentence and translation on a continuous interval, rather than using e.g. binary classification, allowing for more fine-grained selection of translation candidates. 
% In the second round of experiments, we fine-tune a single, multilingual sentiment model to aid in translating from English to Indonesian, and perform another human evaluation. 
The results of human evaluations show that, in comparison to the open-source MT baseline model on top of which our sentiment-based pipeline is built, our pipeline produces more accurate translations of colloquial, sentiment-heavy source texts.\footnote{Code and reference materials are available at \url{https://github.com/AlexJonesNLP/SentimentMT}}.

\end{abstract}

\section{Introduction}
The Web, widespread internet access, and social media have transformed the way people create, consume, and share content, resulting in the proliferation of user-generated content (UGC). UGC—such as social media posts, comments, and reviews—has proven to be of paramount importance both for users and organizations/institutions \cite{pozzi2016sentiment}. As users enjoy the freedoms of sharing their opinions in this relatively unconstrained environment, corporations can analyze user sentiments and extract insights for their decision making processes, \cite{timoshenko2019identifying} or translate UGC to other languages to widen the company's scope and impact. For example, \citet{hale2016user} shows that translating UGC between certain language pairs has beneficial effects on the overall ratings customers gave to attractions and shows on TripAdvisor, while the absence of translation hurts ratings. However, translating UGC comes with its own challenges that differ from those of translating well-formed documents like news articles. UGC is shorter and noisier, characterized by idiomatic and colloquial expressions \cite{pozzi2016sentiment}. Translating idiomatic expressions is hard, as they often convey figurative meaning that cannot be reconstructed from the meaning of their parts \cite{wasow1983idioms}, and remains one of the open challenges in machine translation (MT) \cite{fadaee2018examining}. Idiomatic expressions, however, typically carry an additional property: they imply an affective stance rather than a neutral one \cite{wasow1983idioms}. The sentiment of an idiomatic expression, therefore, can be a useful signal for translation. In this paper, we hypothesize that a good translation of an idiomatic text, such as those prevalent in UGC, should be one that retains its underlying sentiment, and explore the use of textual sentiment analysis to improve translations. 

% Though our paper centers around generating better translations of particular types of texts i.e., UGC, the models we construct are the ones used for sentiment analysis, and our contribution lies in the direct application of these models to the translation process. This paper focuses on sentiment analysis in non-English languages, and how automatic sentiment scoring can be incorporated into the Neural MT (NMT) pipeline. 

Our motivation behind adding sentiment analysis model(s) to the NMT pipeline are several. First, with the sorts of texts prevalent in UGC (namely, idiomatic, sentiment-charged ones), the sentiment of a translated text is often arguably as important as the quality of the translation in other respects, such as adequacy, fluency, grammatical correctness, etc. Second, while a sentiment classifier can be trained particularly well to analyze the sentiment of various texts—including idiomatic expressions \cite{williams2015role}—these idiomatic texts may be difficult for even state-of-the-art (SOTA) MT systems to handle consistently. This can be due to problems such as literal translation of figurative speech, but also to less obvious errors such as truncation (i.e. failing to translate crucial parts of the source sentence). Our assumption however, is that with open-source translation systems such as OPUS MT\footnote{\url{https://github.com/Helsinki-NLP/Opus-MT}}, the correct translation of a sentiment-laden, idiomatic text often lies somewhere lower among the predictions of the MT system, %(i.e. it isn't the “best” candidate as judged by the sentiment-agnostic MT system)
and that the sentiment analysis model can help signal the right translation by re-ranking candidates based on sentiment. 
Our contributions are as follows:
 \setlist{nolistsep}
\begin{itemize}[noitemsep]
\item We explore the idea of choosing translations that minimize source-target sentiment differences on a continuous scale (0-1). Previous works that addressed the integration of sentiment into the MT process have treated this difference as a simple polarity (i.e., positive, negative, or neutral) difference that does not account for the degree of difference between the source text and translation.

%investigate the results of choosing translations that minimize source-target sentiment differences on a continuous (0-1) scale, as opposed to using simple binary or ternary sentiment classification as our metric Unlike with previous papers addressing the integration of sentiment into the MT process, 
\item We focus in particular on idiomatic, sentiment-charged texts sampled from real-world UGC, and show, both through human evaluation and qualitative examples, that our method improves a baseline MT model’s ability to select sentiment-preserving \textit{and} accurate translations in notable cases.
\item We extend our method of using monolingual English and Spanish sentiment classifiers to aid in MT by substituting the classifiers for a single, multilingual sentiment classifier, and analyze the results of this second MT pipeline on the lower-resource English-Indonesian translation, illustrating the generalizability of our approach.
\end{itemize}

\section{Related Work}
%\subsection{Sentiment-oriented MT}
Several papers in recent years have addressed the incorporation of sentiment into the MT process. Perhaps the earliest of these is \citet{sennrich-etal-2016-controlling}, which examined the effects of using honorific marking in training data to help MT systems pick up on the T-V distinction (e.g. informal \textit{tu} vs. formal \textit{vous} in French) that serves to convey formality or familiarity. %\footnote{The distinction some languages have been the informal and formal “you,” present e.g. in French, Spanish, and German} when translating from languages that don’t have it (such as English) to those that do (in this case, German).
\citet{si-etal-2019-sentiment} used sentiment-labeled sentences containing one of a fixed set of sentiment-ambiguous words, as well as valence-sensitive word embeddings for these words, to train models such that users could input the desired sentiment at translation time and receive the translation with the appropriate valence. Lastly, \citet{lohar-2017, lohar-2018} experimented with training sentiment-isolated MT models—that is, MT models trained on only texts that had been pre-categorized into a set number of sentiment classes i.e., positive-only texts or negative-only texts. %, e.g. a model trained using a parallel corpus of only positive-sentiment tweets, or one trained using only negative- and neutral-sentiment tweets, etc. in two different sets of ways using very small amounts We believe our approach is unique in using feature-based candidate selection in regard to sentiment to address this problem, as well as in using precise sentiment scores rather than mere binary or ternary classification to assist the translation process. 
Our approach is novel in using sentiment to re-rank candidate translations of UGC in an MT pipeline and in using precise sentiment scores rather than simple polarity matching to aid the translation process. 

In terms of sentiment analysis models of non-English languages, %It’s worth noting that we are not the first to experiment with sentiment analysis in non-English languages, nor even the first to leverage English-trained sentiment models for analysis in other languages. 
\citet{can-can-can} experimented with using an RNN-based English sentiment model to analyze the sentiment of texts translated into English from other languages, while \citet{balahur-turchi-2012-multilingual} used SMT to generate sentiment training corpora in non-English languages. \citet{Dashtipour:2016} provides an overview and comparison of various techniques used to tackle multilingual sentiment analysis. 

As for MT candidate re-ranking, %We also are not the first to try feature-based reranking methods in the context of MT.
\citet{hadj-ameur} provides an extensive overview of the various features and tools that have been used to aid in the candidate selection process, and also proposes a feature ensemble approach that doesn’t rely on external NLP tools. Others who have used candidate selection or re-ranking to improve MT performance include \citet{shen-etal-2004-discriminative} and \citet{yuan-etal-2016-candidate}. To the best of our knowledge, however, no previous re-ranking methods have used sentiment for re-ranking despite findings that MT often alters sentiment, especially when ambiguous words or figurative language such as metaphors or idioms are present or when the translation exhibits incorrect word order \cite{mohammad2016translation}.

\section{Models and Data}

\subsection{Sentiment Classifiers}
\label{sentiment}
For the first portion of our experiments, we train monolingual sentiment classifiers, one for English and another for Spanish. For the English classifier, %\footnote{In Supplementary Materials: for model parameters, see Sentiment Models $>$ English\_sentiment\_model; for notebook, see Sentiment Classifier Notebooks $>$ English\_sentiment\_notebook.}, 
we fine-tune the BERT Base uncased model ~\citep{devlin-etal-2019-bert}, as it achieves SOTA or nearly SOTA results on various text classification tasks. We construct our BERT-based sentiment classifier model using BERTForSequenceClassification, following %with the aid of a publicly available tutorial 
~\citet{BERT-tutorial}. For our English training and development data, we sample 50K positive and 50K negative tweets from the automatically annotated sentiment corpus described in ~\citet{go-etal-2009} and use 90K tweets for training and the rest for development. For the English test set, we use the human-annotated sentiment corpus also described in ~\citet{go-etal-2009}, which consists of 359 total tweets after neutral-labeled tweets are removed. We use BertTokenizer with ‘bert-base-uncased’ as our vocabulary file and fine-tune a BERT model using one NVIDIA V100 GPU to classify the tweets into positive or negative labels for one epoch using the Adam optimizer with weight decay (AdamW) and a linear learning rate schedule with warmup. We use a batch size of 32, a learning rate of 2e-5, and an epsilon value of 1e-8 for Adam. We experiment with all of the hyperparameters, but find that the model converges very quickly (i.e. additional training after one epoch improves test accuracy negligibly, or causes overfitting). %, and that the other hyperparameters described above were sufficient for obtaining the desired results. 
We achieve an accuracy of 85.2\% on the English test set. %which, though likely not the best result achievable on this test set, was suitable for our purposes, keeping in mind that human inter-annotator agreement on sentiment evaluation is far from 100\%, as noted in \citet{mozetic-the-role} and \citet{validating}.

For the Spanish sentiment classifier%\footnote{In Supplementary Materials: for model parameters, see Sentiment Models $>$ Spanish\_sentiment\_model; for notebook, see Sentiment Classifier Notebooks $>$ Spanish\_sentiment\_notebook.}
, we fine-tune XLM-RoBERTa Large, a multilingual language model that has been shown to significantly outperform multilingual BERT (mBERT) on a variety of cross-lingual transfer tasks ~\citep{conneau-etal-2020-unsupervised}, also using one NVIDIA V100 GPU. We construct our XLM-RoBERTa-based sentiment classifier model again following %with the aid of a publicly available tutorial 
~\citet{BERT-tutorial}. %This model is also based on the classifier used in the BERT tutorial mentioned above ~\citep{devlin-etal-2019-bert}. 
The Spanish training and development data were collected from ~\citet{twitter-sentiment}. %in the form of tweet IDs together with their corresponding human-annotated sentiment labels, and hydrated using Twarc~\citep{docnow-twarc}. 
After removing neutral tweets, we obtain roughly % from the training and development sets to obtain $\approx$
27.8K training tweets and %$\approx$
1.5K development tweets. The Spanish test set is a human-annotated sentiment corpus\footnote{\url{https://www.kaggle.com/c/spanish-arilines-tweets-sentiment-analysis}} containing 7.8K tweets, of which we use roughly 3K after removing neutral tweets and %—unlike with the training and development data—
evening out the number of positive and negative tweets. %to prevent against red-herring “majority accuracy” test scores generated by the model always predicting an overrepresented class. 
%For tokenization, 
We use the XLMRobertaTokenizer with vocabulary file ‘xlm-roberta-large’ and fine-tune the XLM-RoBERTa model to classify the tweets into positive or negative labels. The optimizer, epsilon value, number of epochs, learning rate, and batch size are the same as those of the English model,  %which were once again 
determined via experimentation (without grid search or a more regimented method). %in consideration of the model’s test accuracy.
Unlike with the English model, we found that fine-tuning the Spanish model sometimes produced unreliable results, and so employ multiple random restarts and select the best model, a technique used in the original BERT paper ~\citep{devlin-etal-2019-bert}. The test accuracy on the Spanish model was 77.8\%. %which was again deemed suitable for our purposes in light of the upper bounds on human inter-annotator agreement on sentiment analysis observed in \citet{mozetic-the-role} and \citet{validating}.
\subsection{Baseline MT Models}
\label{baselinemt}
The baseline MT models we use for both English-Spanish and Spanish-English translation are the publicly available Helsinki-NLP/OPUS MT models released by Hugging Face and based on Marian NMT ~\citep{tiedemann-2020, junczys-dowmunt-etal-2018-marian, wolf2019huggingfaces}. Namely, we use both the en-ROMANCE and ROMANCE-en %\footnote{These were the names given to these models by their architects, not by us} 
Transformer-based models, which were both trained using the OPUS dataset~\citep{tiedemann-2017}\footnote{\url{http://opus.nlpl.eu}} with Sentence Piece tokenization and using training procedures and hyperparameters specified on the OPUS MT Github page\footnote{\url{https://github.com/Helsinki-NLP/OPUS-MT-train}} and in ~\citet{tiedemann-2020}. %The en-ROMANCE and ROMANCE-en models we are using have reported respective BLEU scores of 50.1 and 62.2 on the Tatoeba test sets~\citep{tiedemann-2012-parallel}%as per the Hugging Face documentation 
%\footnote{\url{https://huggingface.co/Helsinki-NLP}}. %The Tatoeba test sets are also available on the OPUS MT webpage. Any details not available on the OPUS MT Github page should be found in  ~\citet{tiedemann-2020}. The models we eventually used were derived from a publicly available demo notebook released by Hugging Face \footnote{In Supplementary Materials: see MT Notebooks $>$ En-Romance MT notebook. Also see Romance-En MT notebook and En-Other MT notebook in the same folder}.

\raggedbottom

\section{Method: Sentiment-based Candidate Selection}

\begin{figure}[htbp]
\centerline{\includegraphics[scale=0.6]{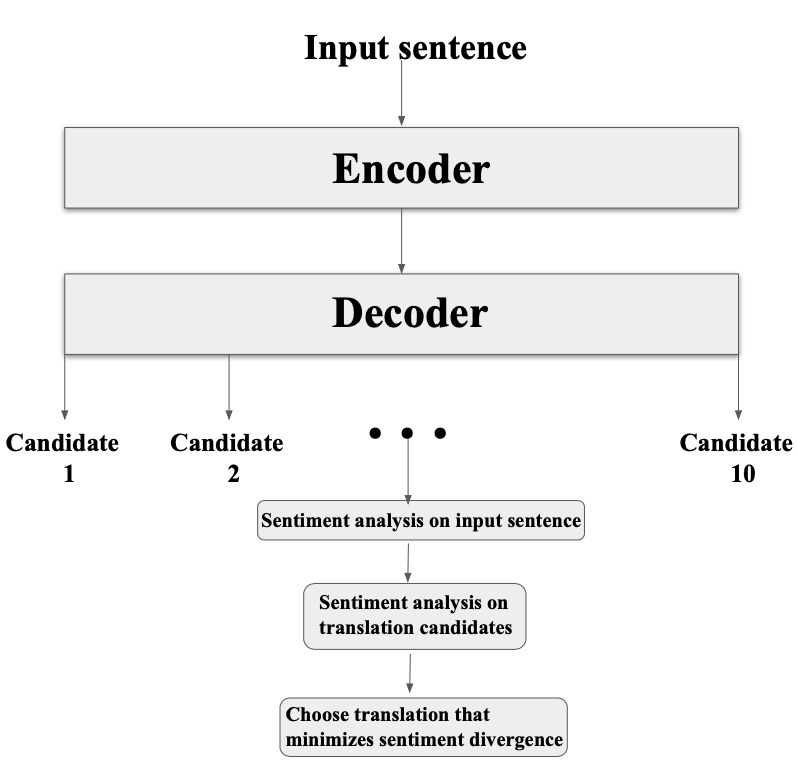}}
\caption{The pipeline for our modified NMT model.}
\vspace{-1.5em}
\label{fig}
\end{figure}

We propose the use of two language-specific sentiment classifiers (which, as we will describe later in the paper, can be reduced to one multilingual sentiment model)—one applied to the input sentence in the source language and another to the candidate translation in the target language—to help an MT system select the candidate translation that diverges the least, in terms of sentiment, from the source sentence (Figure \ref{fig}). 

Using the baseline MT model described in Section \ref{baselinemt}, we first generate $n=10$ best candidate translations using a beam size of 10 at decoding time. We decided on 10 as our candidate number based on the fact that one can expect a relatively low drop off in translation quality with this parameter choice \citep{hasan-etal-2007-large}, while also maintaining a suitably high likelihood of getting variable translations. Additionally, decoding simply becomes too slow in practice beyond a certain beam size.

Once our model generates the 10 candidate translations for a given input sentence, we use the sentiment classifier trained in the appropriate language to score the sentiment of both the input sentence and each of the translations in the interval $[0, 1]$. 
To compute the sentiment score $S(x)$ for an input sentence $x$, we first compute a softmax over the array of logits returned by our sentiment model to get a probability distribution over all $m$ possible classes (here, $m = 2$, since we only used positive- and negative-labeled tweets). Representing the negative and positive classes using the values 0 and 1, respectively, we define $S(x)$ to be the expected value of the class conditioned on $x$, namely $S(x) = \sum_{n=1}^{m} P(c_{n} \mid x)\ v_{n}$, where $c_{i}$ is the $i$th class and $v_{i}$ is the value corresponding to that class. In our case, since we have only two classes and the negative class is represented with value $0$, $S(x) = P(\text{positive class} \mid x)$. After computing the sentiment scores, we take the absolute difference between the input sentence $x$'s score and the candidate translation $t_i$'s score for $i=1, 2,. . . ,10$ to obtain the \textit{sentiment divergence} of each candidate. We select the candidate translation that minimizes the sentiment divergence, namely $y = \argminB_{t_i} |S(t_i)-S(x)|$. %for $i=1, 2,. . . ,10$. %where $T$ is the set of candidate translations. 
Our method of selecting a translation differs from previous works in our use of the proposed sentiment divergence, which takes into account the degree of the sentiment difference (and not just polarity difference) between the input sentence and the candidate translation. 

\section{Experiments\footnote{We make available all our human evaluation data, evaluation questions, responses, and results at \url{https://github.com/AlexJonesNLP/SentimentMT/tree/main/Data\%20and\%20Reference\%20Materials}.}}

\subsection{English-Spanish Evaluation Data}
\label{humanevaldata}

The aim of our human evaluation was to discover how Spanish-English bilingual speakers assess both the quality and the degree of sentiment preservation of our proposed sentiment-sensitive MT model's translations in comparison to those of the human (a professional translator), the baseline MT model (Helsinki-NLP/OPUS MT), and a SOTA MT model, namely Google Translate. 

The human evaluation data %\footnote{In Supplementary Materials: see Translations for Human Evaluation $>$ English-Spanish Translation Sheet.} 
consisted of 30 English (\textit{en}) tweets, each translated using the above four methods to Spanish. We sample 30 English tweets from the English sentiment datasets that we do not use in training (Section \ref{sentiment}) as well as from another English sentiment corpus~\citep{crowdflower-2020}\footnote{\url{https://data.world/crowdflower/apple-twitter-sentiment}}. 
% We sample the 30 Spanish tweets from the Spanish sentiment datasets (Section \ref{sentiment}) as well as from another Spanish sentiment corpus~\citep{spanish-github}\footnote{\url{https://github.com/NatashaSvic/NLP_Spanish_Sentiment_Anaylsis_Text_Generation}}. 
%To construct the English-Spanish translation set, we sample tweets from both the English train and test sets, as well as from another sentiment corpus ~\citep{crowdflower-2020} \footnote{\url{https://rb.gy/cxu4xe}}. To construct the Spanish-English set, we similarly sample tweets from both the Spanish train and test data, as well as from another source ~\citep{spanish-github} \footnote{\url{https://github.com/NatashaSvic/NLP_Spanish_Sentiment_Anaylsis_Text_Generation}}. 
In assembling this evaluation set, we aimed to find a mix of texts that were highly idiomatic and sentiment-loaded—and thus presumably difficult to translate—but also ones that were more neutral in affect, less idiomatic, or some combination of the two. %Though sampling was far from random—which we justify as being in the interest of exhibiting relevant, interesting translations that showcase the possible strengths of our method—we nonetheless try to select samples as well that demonstrated the variability in performance among all three of our MT systems.

\subsection{English-Spanish Evaluation Setup}
\label{spanishsetup}
For the English-Spanish evaluation, we hired two fully bilingual professional translators using contracting site Freelancer \footnote{\url{https://www.freelancer.com/}}. Both evaluators were asked to provide proof of competency in both languages beforehand. The evaluation itself consisted of four translations (one generated by each method: human, baseline, sentiment-MT, Google Translate) for each of the 30 English tweets above, totaling 120 texts to be evaluated. For each of these texts, evaluators were asked to:
\begin{enumerate}
    \item Rate the \textit{accuracy} of the translation on a \textbf{0-5} scale, with 0 being the worst quality and 5 being the best
    \item Rate the \textit{sentiment divergence} of the translation on a \textbf{0-2} scale, with 0 indicating no sentiment change and 2 indicating sentiment reversal
    \item Indicate the reasons for which they believe the sentiment changed in translation
\end{enumerate}
The evaluation instructions and a translation evaluation template are given for reference in the appendix.

\subsection{English-Spanish Evaluation Results}
\label{firstanalysis}

As depicted in Table \ref{tableresult}, the results of the English-Spanish human evaluation show improvements across the board for our modified pipeline over the vanilla baseline model. For the purposes of analysis, we divide the 30 English sentences (120 translations) into two categories: ``all" (consisting of all 120 translations) and ``idiomatic," consisting of 13 sentences (52 translations) deemed particularly idiomatic in nature. Although methods exist for identifying idiomatic texts systematically, e.g. \citet{peng-etal}, we opt to hand-pick idiomatic texts ourselves. We do this in hopes of curating not only texts that contain idiomatic ``multi-word" expressions, but also ones that are idiomatic in less concrete ways, which will enable us to gain more qualitative insights in the evaluation. Examples of such sentences are discussed in Section \ref{discussion}.

In the 'all' subset of the data, we see a +0.12 gain for our modified pipeline over the baseline in terms of accuracy (where higher accuracy is better), as well as a +0.11 improvement in sentiment divergence (where smaller divergence is better). On the idiomatic subset, the differences are more pronounced: we see a +0.80 gain over the baseline for accuracy and a +0.35 improvement in sentiment divergence. While our pipeline lags behind Google Translate in all metrics for English-Spanish—due to the superiority of Google Translate over OPUS MT in multiple regards (training data size, parameters, multilinguality, compute power, etc.)—our modification moves OPUS MT closer to this SOTA system. As a benchmark and to validate the soundness of our evaluation set, we include results for translations performed by a professional human translator, which, as expected, are vastly superior to those for any of the NMT systems used across all metrics and subsets of the data.

We also provide qualitative insights gained from the evaluations, in which evaluators were asked to identify \textit{why} they believe the sentiment of the text \textit{per se} changed in translation. The codes corresponding to these qualitative results are listed in the rightmost column of Table \ref{tableresult}, and may be identified as follows:
\begin{itemize}
\item “MI” indicates the \textsc{m}istranslation of \textsc{i}diomatic/figurative language \textit{per se}
\item “MO” indicates the \textsc{m}istranslation of \textsc{o}ther types of language
\item “IG” indicates \textsc{i}ncorrect \textsc{g}rammatical structure in the translation
\item “IR” indicates \textsc{ir}recoverability of the source text’s meaning, i.e. even the gist of the sentence was gone
\item “LT” indicates a \textsc{l}ack of \textsc{t}ranslatability of the source text to the language in question
\item “O” indicates some \textsc{o}ther reason for sentiment divergence
\end{itemize} 

\begin{table*}[t]%[H]
\centering
\small
\begin{tabular}{ccccccccl}
%\begin{tabularx}{.5\textwidth}{XXXXXX}
 & BLEU & BLEU & BLEU & Accuracy  & SentiDiff  & Accuracy & SentiDiff  & Top-3 Qual. \\
& (Tatoeba) & (all & (idiom. & (all & (all & (idiom.  & (idiom. &  \\
&  & tweets) & tweets) & tweets) & tweets) & tweets) & tweets) &  \\
\textbf{Baseline} &  &  &  &  &  \cr \hline \\
\textit{en}$\rightarrow$es & 31.37 & 38.93 & 39.28 & 2.06 & 0.92 & 1.37 & 1.23 & MI, O, MO \cr \\
% \textit{es}$\rightarrow$en & 4.11 & 1.83 & 4.14 & 1.57 &  MI, MO/IG, IR/LT \cr  \\
% \begin{scriptsize} \textit{en}$\leftrightarrow$es \end{scriptsize} & 3.90 & 1.79 & 3.79 & 1.79 &  MI, MO/IG, IR/LT \cr  \\
%\begin{scriptsize} 
\textit{en}$\rightarrow$id %\end{scriptsize} 
& 31.17 & -- & -- & 2.98 & 0.77 & 2.50 & 1.00 &  MO, O, MI  \cr  \\
\textbf{Modified} &  &  &  &  &  \cr \hline \\
%\begin{scriptsize} 
\textit{en}$\rightarrow$es %\end{scriptsize} 
& 22.15 & 39.10 & 43.47 & 2.18 & 0.81 & 2.17 & 0.88 &  MO, IG, MI  \cr  \\
% \begin{scriptsize} \textit{es}$\rightarrow$en \end{scriptsize} & 2.62 & 2.38 & 4.00 & 0.80 &  MO, MI/IG, LT \cr \\
% \begin{scriptsize} \textit{en}$\leftrightarrow$es \end{scriptsize} & 3.28 & 2.28 & 3.42 & 1.42 &  MO, MI/IG, LT \cr  \\
%\begin{scriptsize} 
\textit{en}$\rightarrow$id %\end{scriptsize} 
& 20.85 & -- & -- & 3.31 & 0.65 & 3.20 & 0.64 &  MO, O, MI \cr  \\
\textbf{Google Transl.} &  &  &  &  &  \cr \hline \\
%\begin{scriptsize} 
\textit{en}$\rightarrow$es %\end{scriptsize} 
& 51.39 & 56.76 & 57.98 & 3.08 & 0.43 & 2.31 & 0.79 &  MI, MO, O \cr  \\
% \begin{scriptsize} \textit{es}$\rightarrow$en \end{scriptsize} & 4.13 & 1.93 & 3.50 & 2.25 &  MI/LT, MO/IG, IR \cr  \\
% \begin{scriptsize} \textit{en}$\leftrightarrow$es \end{scriptsize} & 3.71 & 1.86 & 3.30 & 2.60 &  MI/LT, IG, MO/IR \cr  \\
%\begin{scriptsize} 
\textit{en}$\rightarrow$id %\end{scriptsize} 
& 33.93 & -- & -- & 3.57 & 0.55 & 3.00 & 0.94 &  MO, MI, O/IR  \cr \\
\textbf{Human} &  &  &  &  &  \cr \hline \\
%\begin{scriptsize} 
\textit{en}$\rightarrow$es %\end{scriptsize} 
& 100 & 100 & 100 & 4.28 & 0.10 & 4.44 & 0.08 &  MO, O, IR \cr  \\
% \begin{scriptsize} \textit{es}$\rightarrow$en \end{scriptsize} & 4.69 & 0.92 & 4.33 & 1.33 &  MO/IG/LT, MI/IR  \cr \\
% \begin{scriptsize} \textit{en}$\leftrightarrow$es \end{scriptsize} & 4.63 & 0.89 & 4.56 & 1.56 &  LT, MI/IG/MO, IR  \cr
% \textbf{ALL} &  &  &  &  &  & \hline \\
% \begin{scriptsize} \textit{en}\rightarrow\hspace{-0.2em}es \end{scriptsize} & * & * & * & * & \begin{tiny} M, I/TO, C \end{tiny} & \\
% \begin{scriptsize} \textit{es}\rightarrow\hspace{-0.2em}en \end{scriptsize} & * & * & * & * & \begin{tiny} M, C, I \end{tiny} & \\
% \begin{scriptsize} \textit{en}\leftrightarrow\hspace{-0.2em}es \end{scriptsize} & * & * & * & * & \begin{tiny} M, C, I \end{tiny} & \\
% \begin{scriptsize} \textit{en}\rightarrow\hspace{-0.2em}in \end{scriptsize} & * & * & * & * & \begin{tiny} C, O, M \end{tiny} &
%\end{tabularx}
\end{tabular}
\caption{\label{tableresult} The BLEU scores on the Tatoeba dataset, the accuracy and sentiment divergence scores on Twitter data, and the top 3 reasons given for sentiment divergence for each translation method, language pair, and chosen subset of the Twitter data: all/idiomatic. Note that ratings for each language are given by different sets of evaluators, and shouldn't be compared on a cross-lingual basis.}
%\vspace{-1.5em}
\end{table*}
The top three most frequently cited causes of sentiment divergence for both the baseline and Google Translate were mistranslation of idiomatic language \textit{per se}, mistranslation of other types of language, and other reasons not listed on the evaluation form. For our modified pipeline, the only distinctive top three cause of sentiment divergence was incorrect grammatical structure in the translation; additionally, one human translation was surprisingly flagged as rendering the source text's meaning ``irrecoverable." However, the actual \textit{frequency} of these error codes varied among models. For instance, 'MO' was given 5 times to human translations but 13 times to the baseline model's, and 'O' was given 3 times to Google Translate's translations and 7 times to our pipeline's. Some translations flagged with the '\textbf{O}ther' category are deemed to be of special interest and are discussed in Section \ref{discussion}.

We also noted strong and statistically significant ($p << 0.05$) negative correlations between accuracy and sentiment divergence for both the whole and idiomatic subsets of the data; the values of Pearson's \textit{r} \citep{pearson-coefficient} with their corresponding p-values are reported in Table \ref{pearson}.

Additionally, we measure agreement between the two English-Spanish evaluators using Krippendorff's inter-annotator agreement measure $\alpha$ \citep{krippendorff}, which we choose as a metric in order to compare with previous work examining human agreement on sentiment judgments. In line with \citet{provoost-etal}'s findings of moderate agreement ($\alpha = 0.51$), we see $\alpha$ values ranging from 0.638 to 0.673 for the whole and idiomatic subsets of the data, respectively.

\begin{table}[H]
\scriptsize
\begin{tabularx}{.5\textwidth}{XXX}
& Pearson's \textit{r} (p-value) (all) & Pearson's \textit{r} (p-value) (idiom.) \\
\hline \\
\begin{scriptsize} \textit{en}$\rightarrow$es \end{scriptsize} & -0.764 (3.42e-47) & -0.759 (9.90e-21) \\
\begin{scriptsize} \textit{en}$\rightarrow$id\end{scriptsize} & -0.570 (1.09e-15) & -0.756 (8.67e-14)
\end{tabularx}
\caption{\label{pearson} Pearson's correlation coefficient and corresponding p-value with respect to accuracy and SentiDiff for each of the evaluations, broken down into the full (all) and idiomatic subsets.}
\vspace{-1.5em}
\end{table}

In terms of automatic MT evaluation, we note that although our method causes a decrease in BLEU score on the Tatoeba test data for both languages (Table  \ref{tableresult}: Modified vs. Baseline)—which is to be expected, as Tatoeba consists of ``general" texts as opposed to UGC, and we select potentially non-optimal candidates during re-ranking—our method \textit{improves} over the baseline for the Spanish tweets (and more so on the idiomatic tweets) on which the human evaluation was conducted. This result supports the efficacy of our model in the context of highly-idiomatic, affective UGC, and highlights the different challenges that UGC presents in comparison to more ``formal" text.

Google Translate still outperforms the baseline and our method in terms of BLEU score on Tatoeba and the tweets. The explanation here is simply that the baseline model is not SOTA, which is to be expected given it's a free, flexible, open-source system. However, as our pipeline is orthogonal to any MT model, including SOTA, it could be used to improve a SOTA MT model for UGC.

\section{Method Extension}

\subsection{Translation with Multilingual Sentiment Classifier}

As highlighted in \citet{hadj-ameur}, one of the major criticisms of %feature-based 
decoder-side re-ranking approaches for MT is their reliance on language-specific external NLP tools, such as the sentiment classifiers described in Section \ref{sentiment}. To address the issue of language specificity and to develop a sentiment analysis model that can be used in tandem with MT between any two languages, we develop a multilingual sentiment classifier following \citet{guessing-sentiment}. %used in the main part of our paper. As such, we wondered whether we could address the issue of language specificity with a multilingual sentiment classifier that could be used in tandem with any of the MT models now publicly available on Hugging Face. Inspired by a pioneering blog post on Medium \citep{guessing-sentiment}, we use XLM-RoBERTa to do exactly that. 
Specifically, we fine-tune the XLM-RoBERTa model %(XLMRobertaForSequenceClassification) 
using the training and development data used to train the English sentiment classifier, and the same tokenizer, vocabulary file, hyperparameters, and compute resources (GPU) used in training the Spanish classifier.
We then use this multilingual language model fine-tuned on English sentiment data to perform zero-shot sentiment classification on various languages, and incorporate it into our beam search candidate selection pipeline for MT.

We test the model using the same test data used previously. %employed with both the English and Spanish sentiment classifiers. 
On the English test data, this multilingual model achieves an accuracy of 83.8\%, comparable to the accuracy score achieved using the BERT monolingual model (85.2\%). On the Spanish test set, the multilingual model achieves a somewhat lower score of 73.6\% (\textit{cf.} 77.8\% for the monolingual trained model), perhaps showing the limitations of this massively multilingual model on performing zero-shot downstream tasks.

\subsection{English-Indonesian Evaluation Setup}
\label{secondhumaneval}
We use the multilingual sentiment classifier in our sentiment-sensitive MT pipeline (Figure \ref{fig}) to perform translations on a handful of languages; %shallow qualitative tests on a handful of languages; 
examples from this experimentation are displayed in Tables 4 and 5 in the appendix. 

We perform another human evaluation, this time involving %using 
English$\rightarrow$Indonesian translations in place of English$\rightarrow$Spanish. We choose Indonesian, as it is %\footnote{In Supplementary Materials: see Translations for Human Evaluation $>$ English-Indonesian Survey Translation Sheet.}, as 
a medium-resource language (unlike Spanish, which is high-resource) ~\cite{joshi2020state}, and because we were able to obtain two truly bilingual annotators for this language pair.

The setup of the evaluation %\footnote{In Supplementary Materials: see Survey Templates $>$ English-Indonesian Survey Template.} given to the annotators 
essentially mirrors that of the \textit{en}$\rightarrow$\textit{es} evaluation, except we don't obtain professional human translations as a benchmark for Indonesian, due to the difficulty of obtaining the quality of translation required. Thus, the resulting evaluation set contains only $30*3=90$ translations instead of 120.
%we don’t obtain human translations of the English texts into Indonesian due to time and budget constraints. Second, 
% each annotator is presented with a source text followed by all three translations (baseline, modified, and Google Translate) in random order, as opposed to the first evaluation, where each annotator was given only one translation per source text. This is to allow direct comparison of all translation varieties for each source sentence. %We did this to see whether allowing direct comparison of translations would lead to more consistent intra-rater assessments. Third, 
% Second, we have each annotator assess all of the translations. Since we use the same 30 English tweets as in the first evaluation, each annotator assessed 90 translations. %Note that for both English-Spanish and English-Indonesian evaluation data, the annotators were never shown which translation comes from which model (blind evaluation). 

\subsection{English-Indonesian Evaluation Results}
\label{secondresult}
%The only key issue with the English-Indonesian Evaluation data %\footnote{In Supplementary Materials: see Human Evaluation Results $>$ English-Indonesian Survey Results.} from these surveys 
%was that some sentences were %, for whatever reason, simply 
%not translated by the baseline MT model, possibly because the baseline MT for Indonesian was trained on much less data, thus has lower performance, than the Spanish model. % Helsinki-NLP/OPUS MT model (or at least certain candidates weren’t). Even though this indicates an issue with this MT system, we discarded these results due to their lack of informativity about the influence of SD on translation quality. 
The accuracy and sentiment divergence averages for different subsets of the \textit{en}-\textit{id} data are located in Table \ref{tableresult}, and we direct readers to Section \ref{firstanalysis} for a qualitative discussion of these results. %We make a couple of observations about the results in Table 1: one, that our modified model outperforms the baseline in every single category, strongly demonstrating the advantages of our method on this language pair and direction; and two, that while our model lags behind the SOTA on both general accuracy and SD, it beats the SOTA on both these measurements on the idiomatic set, again displaying the strengths of our method. \\
Quantitatively, we observe that our modified model outperforms the baseline in accuracy and sentiment divergence on every subset of the \textit{en}-\textit{id} data, while being comparable or better than Google Translate on the ``all" and idiomatic subsets, respectively (Table \ref{tableresult}). Specifically, on the ``all" subset we see improvements of +0.33 and +0.12 over the baseline for accuracy and sentiment divergence, respectively, and on the idiomatic subset we see respective improvements of +0.70 and +0.36. Google Translate achieves slightly better accuracy and sentiment preservation overall (+0.26 and +0.10 over our pipeline for accuracy and sentiment divergence, respectively), but lags behind our pipeline in the idiomatic category (-0.20 and -0.30 for accuracy and sentiment divergence, respectively, compared to our pipeline).

Qualitatively, we see very similar reasons listed for sentiment divergence as we did for English-Spanish: each of the NMT systems we looked at had errors most frequently in the MI, MO, and O categories, denoting mistranslation of idiomatic language, mistranslation of other types of language, and other reasons for sentiment divergence, respectively; with MO being more frequent than MI in English-Indonesian evaluations, potentially due to lower MT performances for this language than Spanish (i.e., BLEU score for English-Indonesian modified model is 20.85 on the Tatoeba dataset compared to 22.15 for English-Spanish%\footnote{https://huggingface.co/Helsinki-NLP/opus-mt-en-id}
). 
However, as noted in the analysis of the previous evaluation, not all of these errors occurred with equal frequency across systems. For instance, Google Translate and the human translator produced less errors overall than the OPUS MT system, so the error codes should be interpreted as indicating the relative frequency and prevalence of certain translation errors that affect sentiment, not as markers to be compared on a system-to-system basis. As with the English-Spanish evaluation, certain qualitative observations made by our evaluators will be discussed further in Section \ref{discussion}. 
In line with results on the previous evaluation, accuracy and sentiment divergence are shown to be strongly negatively correlated, with Pearson's \textit{r} values of -0.570 and -0.756 for the whole and idiomatic subsets of the data, respectively, both of which are statistically significant ($p << 0.05$) and are displayed in Table \ref{pearson}.
%, which demonstrates the strengths of our proposed method on this language pair and direction.  

% Qualitatively, we see different reasons given for sentiment divergence with this \textit{en}-\textit{id} data as a whole compared to the \textit{en}-\textit{es}, with non-idiomatic mistranslation (cited 50 times), other reasons (37), and idiomatic mistranslation (36) being most frequent. The swapping of non-idiomatic and idiomatic mistranslation may be due to the generally higher accuracy of \textit{en}$\leftrightarrow$\textit{es} MT compared to \textit{en}$\leftrightarrow$\textit{id} MT, with more non-idiomatic words being mistranslated in the second case.

% Among the “other” reasons cited, the most frequent were missing words (12) and non-translation of idiomatic words/acronyms (9). Both of these issues are predictable given the state of \textit{en}$\leftrightarrow$\textit{id} MT.

\begin{table}[H]
\scriptsize
\begin{tabularx}{.5\textwidth}{XXXXX}
& %$\kappa$\&
%$\alpha$ 
acc. (all) & %$\kappa$\&
%$\alpha$ 
SentiDiff (all) & %$\kappa$\&
%$\alpha$ (
acc (idiom.) & %$\kappa$\&
%$\alpha$ (
SentiDiff (idiom.) \\
\hline \\
\textit{en}$\rightarrow$es  & %0.273\&
0.675 & %0.212\&
0.638 & %0.226\&
0.767 & %0.256\&
0.673 \\
\textit{en}$\rightarrow$id  & %0.273\&
0.661 & %0.212\&
0.516 & %0.226\&
0.612 & %0.256\&
0.541
\end{tabularx}
\caption{\label{agreement} Values of %Cohen's kappa agreement measure (denoted $\kappa$) and
Krippendorff's alpha agreement measure %reliability measure 
$\alpha$ for both sets of evaluations with respect to accuracy (``acc.") and sentiment divergence (``SentiDiff") across different subsets.}
\vspace{-1.5em}
\end{table}

%Additionally, Table 2 shows the values of Pearson’s \textit{r} for the full and idiomatic subsets of the data (the idiomatic English texts were the same as in the first evaluation), while 
Table \ref{agreement} shows %Cohen’s kappa agreement measure \citep{cohen-kappa} and 
Krippendorff’s alpha agreement measure %reliability measure 
\citep{krippendorff} for accuracy and sentiment divergence across both subsets, indicating moderate agreement, %rates between annotators, 
with higher agreement on accuracy. As was found with the English-Spanish evaluation, this is in line with previous findings of moderate human agreement on sentiment judgement (Krippendorff's $\alpha$=0.51) \citep{provoost-etal}.  %We note a relatively weak negative correlation for accuracy-SD on the full dataset (\textit{r}=-0.570), contrasted with a much stronger one on the idiomatic set (\textit{r} = -0.756), which makes sense as the idiomatic texts were selected for their sentiment-bearing nature. We note that the Cohen’s kappa and Krippendorff’s alpha ratings are quite low across all subsets, indicating very imperfect agreement between the two annotators. However, while such ratings may indicate poor-quality data in other settings, we speculate that the inherent ambiguity of translation accuracy and SD assessment may make these ratings statistically acceptable in this context.

\section{Discussion}
\label{discussion}
Our experimentation with the various MT models generated a number of interesting example cases concerning the translation of idiomatic language. %We saw tweets in which our model succeeded compared to baseline and Google Translate. 
For example, given the tweet ``Time Warner Road Runner customer support here absolutely blows," the baseline MT gives a literal translation of the word ``blows" as ``pukulan" (literally, ``hits") in Indonesian; Google Translate gives a translation ``hebat" (``awesome") that is opposite in sentiment to the idiomatic sense of the word ``blows" (``sucks") in English; and our model gives a translation closest in meaning and sentiment to ``blows," namely ``kacau" (approx. ``messed up" in Indonesian). There are also cases where our model gives a translation that is closer in degree of sentiment than what Google Translate produces. Given the source text ``Yo @Apple fix your shitty iMessage," Google Translate produces ``Yo @Apple perbaiki iMessage \textit{buruk} Anda" (``Yo @Apple fix your \textit{bad} iMessage"), which has roughly the same polarity as the source tweet. By contrast, our proposed model produces ``Yo @Apple perbaiki imessage \textit{menyebalkan} Anda," using the word ``menyebalkan" (``annoying") instead of ``buruk," which conveys a closer sentiment to ``shitty" than simply ``bad". 

In general, we observe that translation of idiomatic language commonly results in literal translation of the language by all MT models, even for a SOTA MT system like Google Translate. For example, given tweets like ``Okay, first assessment of the Kindle... it fucking rocks!," all the MT models translate ``rocks" to ``batu" (meaning ``stones") in Indonesian. 
There are also cases where the tweet simply can't be translated into the language without knowledge of its larger context, e.g., ``I'm sorry—I'm feeling kinda yucky myself—5am is going to come too quick." Here ``yucky" can only be translated to ``jijik" (``disgusting") in Indonesian, which does not capture precisely the sense of ``yucky" here as ``sick" or ``crummy."

In terms of translating sentiment-charged texts in general, there are cases where the baseline MT model translates positive/negative tweets as neutral, in line with previous findings of MT systems' translation of sentiment-charged texts \cite{mohammad2016translation}. In these cases, the modified model helps select the translation with the correct sentiment. For instance, given the tweet ``Went to see the Star Trek movie last night.  Very satisfying," the baseline MT model produces ``Pergi menonton film Star Trek tadi malam." (``Went to watch the Star Trek movie last night.") without the accompanying sentiment. The modified model, on the other hand, translates it to ``Pergi menonton film Star Trek tadi malam. Sangat memuaskan," correctly choosing the translation with the sentiment ``Very satisfying." In these cases, our modified pipeline acts as a sort of \textit{de facto} coverage penalty in decoding, compensating indirectly for an apparent shortcoming of OPUS MT's open source system.

Acronyms such as ``tbh" and ``smh" made for another interesting case, as they weren't translated by any of the MT models for any language pairing, despite their common occurrence in UGC. 
Lastly, there were grammatical problems such as incorrect word order, %made for another interesting case, 
particularly in 
%Grammatical factors made for another interesting case, particularly in 
some \textit{en}$\rightarrow$\textit{id} baseline/modified translations.
, %of the baseline and modified MT, 
where e.g. the subject and object were reversed, which may be due to the lack of publicly available data for training the Indonesian baseline MT model. Such errors sometimes, but not always, had a noticeable effect on the sentiment of the resulting translation, as noted by our evaluators for e.g. English-Spanish (see ``Top-3 Qual." in Table \ref{tableresult} under ``Modified").

The evaluators of the English-Spanish translations provided us with rich qualitative commentary as well. For example, for the sentence ``Okay, first assessment of the Kindle . . . it fucking rocks!," which our pipeline translates as ``Bien, la primera evaluación del Kindle... ¡está jodidamente bien!," one evaluator notes that ``The word `bien' has less positive sentiment than the word `genial' (which is a closer translation for   `it rocks')." For the sentence ``Just broke my 3rd charger of the month. Get your shit together @apple," which is translated by the professional translator as ``Se acaba de romper mi tercer cargador del mes. Sean más eficientes @apple," one evaluator acutely notes that ``The expression `Get your shit together' was translated in a more formal way (it loses the vulgarism). I would have translated it as `Poneos las pilas, joder' to keep the same sentiment. We could say that this translation has a different diaphasic variation than the source text." This demonstrates that sentiment preservation is a problem not only for NMT systems, but for human translators as well. The same evaluator also notes that ``The acronym `tbh' was not translated" in the sentence ``@Apple tbh annoyed with Apple's shit at the moment," and says ``this acronym is important for the sentiment because it expresses the modality of the speaker." In another example, we see our sentiment-sensitive pipeline helping the baseline distinguish between such a semantically fine-grained distinction as that between ``hope" and ``wish": the baseline translates the sentence ``@Iberia Ojalá que encuentres pronto tuequipaje!!" as ``@Iberia I \textit{wish} you’d find your luggage soon!!," while our pipeline correctly chooses ``@Iberia  I  \textit{hope} you will find your luggage soon!!." Clearly, we see considerable overlap in the causes of sentiment divergence between Spanish and Indonesian, despite the fact that these are typologically disparate languages with differing resource capacities in regard to MT.

In other languages, similar patterns of idiomatic mistranslation are observed. In French, for example, both Google Translate and the baseline model translate ``Why are people such wankers these days?" as ``Pourquoi les gens sont-ils si \textit{branleurs} ces jours-ci?," while our pipeline produces ``Pourquoi les gens sont-ils si \textit{cons} ces jours-ci?." In this case, the use of ``si" before the noun ``branleurs" doesn't sound idiomatic (``\textit{tels} branleurs" would fit better), but the degree term \textit{si} makes sense before adjective ``cons" (approx. ``stupid" or ``irritating"), which is also a more versatile, common, and idiomatic term in French than ``branleurs." This is a prime example of a source text for which the most idiomatic translation isn't the most literal one, even though the source text doesn't contain figurative language \textit{per se}. The baseline system evidently benefits from a sentiment-sensitive step in the decoder that nudges it toward this choice without being specifically trained to reward idiomaticity.

In terms of automatic MT evaluation, %we note that although our method causes a decrease in BLEU score on the Tatoeba test data for both languages—which is to be expected, as Tatoeba consists of "general" texts as opposed to UGC, and we select potentially non-optimal candidates during re-ranking—
our method improves over the baseline for the Spanish tweets on which the human evaluation was conducted. This result supports the efficacy of our model in the context of highly-idiomatic, affective UGC. %, and highlights the different challenges that UGC presents in comparison to more "formal" text.
And while Google Translate still outperforms the baseline and our pipeline in terms of BLEU score on Tatoeba (for both languages) and the tweets (for which only Spanish had a gold-standard benchmark)--given that the baseline model that we built our pipeline on is not SOTA--our pipeline can be added to any MT system and can also improve SOTA MT for UGC. 

%The explanation here is simply that the baseline model is not SOTA, which is to be expected given it's a free, flexible, open-source system. 
Furthermore, %while our MT pipeline isn't topping any leaderboards, 
our approach also lends itself to many practical scenarios, e.g. companies who are interested in producing sentiment-preserving translations of large bodies of UGC but who lack the sufficient funds to use a subscription API like Google Cloud Translation. In these contexts, it may be beneficial—or even necessary—to improve free, open-source software in a way that is tailored to one's particular use case (thus the idea of ``customized MT" that many companies now offer), instead of opting for the SOTA but more costly software. 

More generally, since our approach shows that we can improve performance of an MT model for a particular use case i.e., UGC translation using signals beyond translation data that is relevant for the task at hand i.e., sentiment, it will be interesting to explore other signals that are relevant for improving MT performance in other use cases. It will also be interesting to explore the addition of these signals in a pipeline (our current method), as implicit feedback such as in \cite{wijaya2017learning}, or as explicit feedback in an end-to-end MT model for example, as additional loss terms in supervised \cite{wu2016google}, weakly-supervised \cite{kuwanto2021low}, or unsupervised \cite{artetxe2017unsupervised} MT models. 

Beyond the potential engineering contribution for low-resource, budget-constrained settings, our experiments also offer rich qualitative insights regarding the causes of sentiment change in (machine) translation, opening up avenues to more disciplined efforts in mitigating and exploring these problems. 

% Our MT-pipeline-based experiments with both the language-specific sentiment classifiers and the multilingual classifier confirmed our assumptions that there are certain types of texts—in particular, highly idiomatic ones—on which SOTA MT systems often fail, but that, if bolstered by an external sentiment tool, can be successfully and idiomatically translated to other languages. Though the quantitative results from our human evaluation were equivocal at best, and dubious at worst, we nonetheless find our qualitative results—i.e. the translations displayed in Tables 1 and 2, plus those available at various links in the footnotes—to be revealing and promising. \\
% We anticipate one criticism being that, instead of using an external NLP tool to aid in translating these particular sorts of idiomatic texts—ones found frequently on social media, in product reviews, etc.—we could instead train an MT system using a parallel corpus / parallel corpora from these domains. The problems with this approach are, however, evident; among them are (1) the cost and difficulty of obtaining such corpora; (2) the difficulty of scaling this approach; and (3) the domain-specificity and linguistic (e.g. lexical) limitations of such a model, particularly if it isn’t also trained on out-of-domain data. However, we welcome other researchers to try this approach and others and compare the results to ours.

\section{Conclusion}
\label{conclusion}

% While many of the greatest advancements in MT have, in recent years, taken the form of high-cost, high-resource, high-sophistication approaches—the application of RNNs, different types of attention mechanisms, architectural modifications, massive increases in data, etc.—there remain cases of texts that, however common and easy for humans to understand, continue to stymie even SOTA MT systems. While some of these problems may eventually fall to advances of the sort described above, we need in the short term to wield creative, low-cost NLP strategies to our advantage; as such, 
%In conclusion, 
In this paper we use sentiment analysis models %SOTA language models %—the experimentation on which is still in its nascent stages—
to help publicly available MT baseline models select sentiment-preserving translations. Conducting both automated and manual evaluations, we show that our architecturally simple, low-resource, and generalizable approach can produce surprisingly good results for translating the sort of idiomatic texts prevalent in UGC. In the future, we will explore the use of sentiment as additional feedback for training an MT model end-to-end. 

% \section{Citations repo (DELETE before submission)}

% ~\citep{balahur-turchi-2012-multilingual}
% ~\citep{conneau-etal-2020-unsupervised}
% ~\citep{can-can-can}
% ~\citep{crowdflower-2020}
% ~\citep{Dashtipour:2016}
% ~\citep{devlin-etal-2019-bert}
% ~\citep{docnow-twarc}
% ~\citep{go-etal-2009}
% ~\citep{hadj-ameur}
% ~\citep{junczys-dowmunt-etal-2018-marian}
% %~\citep{pearson-coefficient}
% ~\citep{lohar-2017}
% ~\citep{lohar-2018}
% ~\citep{BERT-tutorial}
% ~\citep{guessing-sentiment}
% ~\citep{twitter-sentiment}
% ~\citep{spanish-github}
% ~\citep{sennrich-etal-2016-controlling}
% ~\citep{shen-etal-2004-discriminative}
% ~\citep{si-etal-2019-sentiment}
% ~\citep{tiedemann-2012-parallel}
% ~\citep{tiedemann-2017}
% ~\citep{tiedemann-2020}
% ~\citep{kaggle-spanish-data}
% ~\citep{attention-is-all-you-need}
% ~\citep{wolf2019huggingfaces}
% ~\citep{yuan-etal-2016-candidate}

\section*{Acknowledgements}

We would like to thank Boston University for their generous funding of this project. We also appreciate the work of Chris McCormick and Nick Ryan in constructing the BERT tutorial we used as a blueprint for all three sentiment models.
% , and extend our gratitude as well to the Hugging Face team, the Marian team, and the Helsinki Language Technology Research Group for their efforts in making NLP models publicly available for researchers.

\bibliography{submission_bib}
\bibliographystyle{acl_natbib}

\raggedbottom

\onecolumn

\appendix

\section*{Appendix}

\section{Example Translations}

\newcolumntype{b}{>{\hsize=1.7\hsize}X}
\newcolumntype{s}{>{\hsize=.3\hsize}X}

\begin{table}[H]
\small
\centering
\renewcommand{\arraystretch}{1.2}
\begin{tabularx}{0.8\textwidth}{sb}
%\normalsize \textbf{Text Type} & \normalsize \textbf{Text} \\
\hspace{-1em} \textbf{French} \\ \toprule
Original & \textcolor{red}{Yo} @Apple fix your \textcolor{red}{shitty} iMessage \\ %\hline
Baseline & \textcolor{red}{Yo} @Apple répare ta \textcolor{red}{merde} iMessage \\ %\hline
Modified & \textcolor{red}{Yo} @Apple réparer votre \textcolor{red}{sale} iMessage \\ \hline
% Google Translate & Yo @Apple corrige ton iMessage meridique \\ \hline \hline
Original & Why are people \textcolor{red}{such wankers these days}? \\ %\hline
Baseline & Pourquoi les gens sont-ils \textcolor{red}{si branleurs ces jours-ci}? \\ %\hline
Modified & Pourquoi les gens sont-ils \textcolor{red}{si cons ces jours-ci}? \\ %\hline %\hline
%\bottomrule
% Google Translate & Pourquoi les gens sont-ils si branleurs ces jours-ci? \\ \hline \hline
\\
\hspace{-1em} \textbf{Italian} \\ \toprule
Original & \textcolor{red}{Yo} @Apple fix your \textcolor{red}{shitty} iMessage \\ 
Baseline & \textcolor{red}{Yo} @Apple riparare la tua \textcolor{red}{merda} iMessage \\ 
Modified & \textcolor{red}{Yo} @Apple riparare il vostro iMessaggio \textcolor{red}{di merda} \\ 
%\bottomrule
% Google Translate & Yo @Apple aggiusta il tuo iMessage di merda \\ \hline \hline
\\
\hspace{-1em} \textbf{Portuguese}\\
\toprule
Original & Time Warner Road Runner customer support here \textcolor{red}{absolutely blows}. \\
Baseline & O suporte ao cliente do Time Warner Road Runner é \textcolor{red}{absolutamente insuportável}. \\
Modified & O suporte ao cliente do Time Warner Road Runner aqui é \textcolor{red}{absolutamente estragado}. \\ \\ %\bottomrule \\
% Google Translate & O suporte ao cliente da Time Warner aqui é absolutamente incrível \\ \hline \hline
% \end{tabularx}
% \end{table}

% \begin{table}
% \tiny
% \begin{tabularx}{.48\textwidth}{XX}
\hspace{-1em} \textbf{Indonesian} \\ \toprule
Original & Time Warner Road Runner customer support here \textcolor{red}{absolutely blows}. \\
Baseline & Dukungan pelanggan Pelarian Jalan Warner waktu \textcolor{red}{di sini benar-benar pukulan}. \\
Modified & Dukungan pelanggan \textcolor{red}{di sini benar-benar kacau}. \\ \hline
% Google Translate & Dukungan pelanggan Time Warner di sini benar-benar hebat. \\ \hline \hline
Original & \textcolor{red}{Yo} @Apple fix your \textcolor{red}{shitty} iMessage \\
Baseline & \textcolor{red}{Yo} @Apple perbaiki pesan \textcolor{red}{menyebalkanmu} \\
Modified & \textcolor{red}{Yo} @Apple perbaiki imessage \textcolor{red}{menyebalkan} Anda \\ \hline
Original & Went to see the Star Trek movie last night.  \textcolor{red}{Very satisfying}. \\
Baseline & Pergi menonton film Star Trek tadi malam. \\
Modified & Pergi menonton film Star Trek tadi malam. \textcolor{red}{Sangat memuaskan}. \\
% Google Translate & Yo @Apple perbaiki iMessage buruk Anda \\ \hline \hline
\\
\hspace{-1em} \textbf{Croatian} \\ \toprule
Original & Time Warner Road Runner customer support here \textcolor{red}{absolutely blows}. \\
Baseline & Podrška za klijente Time Warner Road Runnera ovdje \textcolor{red}{apsolutno udara}. \\
Modified & Podpora klijenta Time Warner Road Runnera ovdje je \textcolor{red}{apsolutno pukla}. \\
% Google Translate & Time korisnička podrška Time Warner Road Runner apsolutno puše. \\ \hline \hline
\\
\hspace{-1em} \textbf{German} \\ \toprule
Original & I need a new boyfriend... I'm \textcolor{red}{stuck in a rut}. \\
Baseline & Ich brauche einen neuen Freund. \\
Modified & Ich brauche einen neuen Freund, ich \textcolor{red}{stecke in einer Falle}. \\ \hline
% Google Translate & Ich brauche einen neuen Freund ... Ich stecke in der Brunft. \\ \hline \hline
Original & \textcolor{red}{aww that stinks} \\
Baseline & \textcolor{red}{Das stinkt!} \\
Modified & \textcolor{red}{Oh, das stinkt doch!} \\ \\
% Google Translate & aww das stinkt \\ \hline \hline
% \end{tabularx}
% \end{table}

\hspace{-1em} \textbf{Finnish} \\ \toprule
Original & Could Time Warner Cable \textcolor{red}{suck more}? \\
Baseline & Voisiko Time Warner Cable \textcolor{red}{imeä enemmän}? \\
Modified & Voisiko Time Warner Cable \textcolor{red}{olla enemmän surkea}? \\ \hline
% Google Translate & Voisiko Time Warner -kaapeli imeä enemmän \\ \hline \hline
% Google Translate & Tarvitsen uuden poikaystävän ... Olen jumissa rutista. \\ \hline \hline
Original & I'm sorry—I'm \textcolor{red}{feeling kinda yucky} myself—5am is going \textcolor{red}{to come too quick}. \\
Baseline & Olen pahoillani, olen \textcolor{red}{itsekin aika naljaillen}, että aamuviideltä \textcolor{red}{tulee liian nopeasti}. \\
Modified & Olen pahoillani, \textcolor{red}{että olen itse vähän kuvottava}, mutta aamuviideltä \textcolor{red}{tulee liian nopea}. \\
% Google Translate & Olen pahoillani - Minusta tuntuu siltä, että sinulla on ikävyyttä - 5:00 tulee liian nopeasti. \\ \hline \hline
\end{tabularx}
\end{table}

\begin{table}[ht!]
\small
\centering
\renewcommand{\arraystretch}{1.2}
\begin{tabularx}{0.8\textwidth}{sb}
\hspace{-1em} \scriptsize \textbf{English-Spanish} \\ \toprule
Original & Time Warner Road Runner customer support here \textcolor{red}{absolutely blows}. \\
Baseline & Time Warner Road Runner atención al cliente aquí \textcolor{red}{absolutamente golpes}. \\
Modified & El servicio de atención al cliente de Warner Road Runner aquí \textcolor{red}{es absolutamente malo}. \\ \hline
% Google Translate & La atención al cliente de Time Warner Road Runner aquí es absolutamente increíble. \\ \hline
% Human & El servicio de atención al cliente de Warner Road Runner es malísimo aquí. \\ \hline \hline
% Original & Okay, first assessment of the Kindle . . . it fucking rocks! \\ \hline
% Baseline & De acuerdo, la primera evaluación del Kindle... ¡es una maldita piedra! \\ \hline
% Modified & Bien, la primera evaluación del Kindle... ¡está jodidamente bien! \\ \hline
% Google Translate & Bien, primera evaluación del Kindle. . . es jodidamente genial! \\ \hline
% Human & De acuerdo, la primera evaluación del Kindle... ¡Es lo máximo! \\ \hline \hline
Original & Could Time Warner Cable \textcolor{red}{suck more}? \\
Baseline & ¿Podría Time Warner Cable \textcolor{red}{chupar más}? \\
Modified & ¿Podría Time Warner Cable \textcolor{red}{apestar más}? \\ \hline
Original & Went to see the Star Trek movie last night. \textcolor{red}{Very satisfying}. \\
Baseline & Fui a ver la película de Star Trek anoche. \\
Modified & Fui a ver la película de Star Trek anoche, \textcolor{red}{muy satisfactoria}. \\
% Google Translate & ¿Podría Time Warner Cable chupar más? \\ \hline
% Human & ¿Podría Time Warner Cable apestar más? \\ \hline \hline
\\
\hspace{-1em} \scriptsize \textbf{Spanish-English} \\ \toprule
Original & Vivo con el miedo que Ryanair me cancelen los vuelos...  \textcolor{red}{Que poquito me fio de ellos} \\
Baseline & I live with the fear that Ryanair will cancel my flights... \textcolor{red}{that I don't give a damn about them}. \\
Modified & I live with the fear that Ryanair will cancel my flights... \textcolor{red}{what little do I trust them with}? \\ \hline
% Google Translate & I live with the fear that Ryanair will cancel my flights ... how little I trust them \\ \hline
% Human & I always have the constant fear that Ryanair will cancel my flights…I don’t trust them at all!!! \\ \hline \hline
% Original & @Iberia Viajo pronto con ustedes !!! Ojalá gane !!! \\ \hline
% Baseline & @Iberia I travel soon with you!!! I wish I won!!! \\ \hline
% Modified & @Iberia I travel soon with you!!! I hope I win!!! \\ \hline
% Google Translate & @Iberia I travel soon with you !!! I hope I win !!! \\ \hline
% Human & @Iberia I travel with you soon!!! I hope I won!!! \\ \hline \hline
Original & La verdad es que a día de hoy, con Ryanair las cosas están \textcolor{red}{muy baratas para viajar}. \\
Baseline & The truth is that nowadays, with Ryanair things are \textcolor{red}{too cheap to travel}. \\
Modified & The truth is that today, with Ryanair, things are \textcolor{red}{very cheap to travel}. \\ \hline
Original & @Iberia \textcolor{red}{Ojalá} que encuentres pronto tu equipaje!! \\
Baseline & @Iberia I \textcolor{red}{wish} you'd find your luggage soon!! \\
Modified & @Iberia I \textcolor{red}{hope} you will find your luggage soon!! \\ \bottomrule
% Google Translate & The truth is that today, with Ryanair things are very cheap to travel. \\ \hline
% Human & The truth is that today Ryanair’s offers are very tempting when it comes to traveling. \\ \hline \hline
\end{tabularx}
\caption{\label{examples-bi} Example texts exhibiting our pipeline's performance using the sentiment models from the first and second human evaluations.}
\end{table}

\begin{multicols}{2}
\section{Evaluation Instructions}

The following are excerpts from the instructions given to evaluators for both the English-Spanish and English-Indonesian evaluations: \\
\textsl{The document you are now looking at should contain prompts numbered up to 120. For each of these prompts, you will be asked to do three things: 
\\
\begin{enumerate}
\item Rate the \textit{accuracy} of the translation. Please rate the \textbf{accuracy} of the translation on a \textbf{0 to 5} scale, where \textbf{0} indicates an “awful” translation, \textbf{2.5} indicates a “decent” translation, and \textbf{5} indicates a “flawless” translation. Feel free to use half-increments, i.e. 0.5, 1.5, 2.5, 3.5, or 4.5, and go with your gut as a fluent speaker of both languages.
\item Rate the \textit{sentiment divergence} of the translation. The \textbf{sentiment} of a text corresponds to the emotion it conveys. For example, “I am happy” has a very positive sentiment, and “The weather is terrible” has a very negative sentiment. Please rate the \text{sentiment divergence} on a \textbf{0 to 2} scale, where \textbf{0} indicates that the sentiment of the source sentence \textbf{perfectly matches} that of the translation and \textbf{2} indicates that the sentiment of the source sentence is the \textbf{opposite} of that of the translation. As when rating accuracy, feel free to use half-increments, i.e. 0.5, 1.5.
\item Indicate the \textit{reasons} for sentiment divergence. Please \textbf{bold} the statements which you believe accurately describe \textbf{why} you believe the sentiment changed from the source sentence to the translation. For example if the source sentence was “Oh my god, it’s raining cats and dogs!” and the translation was \textit{Ay dios mio, esta lloviendo perros y gatos,} ” you should bold the statement “The translation contained literal translation(s) of figurative English language.”
\end{enumerate}}

\section{Sample Prompt}
Below is an example of a translation evaluation prompt that evaluators were asked to respond to:
\textsl{
\begin{itemize}
\item{Accuracy:}
\item{Sentiment divergence:}
\item{Please ​bold​ all of the below which had an effect on the ​sentiment​ of the translation:}
\begin{enumerate}
    \item The translation contained literal translation(s) of figurative English language
    \item The translation contained other types of mistranslated words
    \item The original (English) sentence can’t be properly translated to Spanish
    \item The overall meaning of the English sentence changed substantially when translated to Spanish (i.e. even the gist of the English sentence is unrecoverable)
    \item The sentiment didn’t change at all
    \item Other (please write in next to this)
\end{enumerate}
\end{itemize}}
\end{multicols}

\end{document}